\documentclass[10pt,twocolumn,letterpaper]{article}

\usepackage{cvpr}
\usepackage{times}
\usepackage{epsfig}
\usepackage{graphicx}
\usepackage{amsmath}
\usepackage{amssymb}
\usepackage{subfigure}
\usepackage{array}

\makeatletter
\newcommand{\thickhline}{%
    \noalign {\ifnum 0=`}\fi \hrule height 1pt
    \futurelet \reserved@a \@xhline
}
\newcolumntype{"}{@{\hskip\tabcolsep\vrule width 1pt\hskip\tabcolsep}}
\makeatother

\usepackage{xcolor}
\usepackage{booktabs}

\usepackage[pagebackref=true,breaklinks=true,letterpaper=true,colorlinks,bookmarks=false]{hyperref}

\cvprfinalcopy 

\def\cvprPaperID{28} 

\ifcvprfinal\fi

\begin{document}

\title{Event-based attention and tracking on neuromorphic hardware}

\author{Alpha Renner\\
Institute of Neuroinformatics\\
University of Zurich and ETH Zurich\\
{\tt\small alpren@ini.uzh.ch}
\and
Matthew  Evanusa\\
University of Maryland, MD, USA\\
{\tt\small matthew.evanusa@gmail.com}
\and
Yulia  Sandamirskaya\\
Institute of Neuroinformatics\\
University of Zurich and ETH Zurich\\
{\tt\small ysandamirskaya@ini.uzh.ch}
}

\maketitle
\thispagestyle{empty}

\begin{abstract}
We present a fully event-driven vision and processing system for selective attention and tracking, realized on a neuromorphic processor Loihi  interfaced to an event-based Dynamic Vision Sensor DAVIS. The attention mechanism is realized as a recurrent spiking neural network that implements attractor-dynamics of dynamic neural fields. We demonstrate capability of the system to create sustained activation that supports object tracking when distractors are present or when the object slows down or stops, reducing the number of generated events. 
\end{abstract}

\section{Introduction}

Event-based sensors send out data packages, or events, from each pixel asynchronously when the pixel detects a local brightness change,  rather than reading all pixels and sending out frames at a constant rate. Such event-based sensing allows us to perform some vision tasks extremely efficiently, reducing the amount of required computation, transmitted data, and power consumption. Many event-based vision pipelines and architectures have been developed over the last decade \cite{Lagorce2015,Lagorce2016}, which  address such vision tasks as stereo vision \cite{OsswaldEtAl2017}, 3D pose estimation \cite{Valeiras2015,CarneiroEtAl2013},  or optical flow \cite{Benosman2012}.  These event-based pipelines are typically implemented on conventional Von Neumann computer architectures. While such implementations try to make the best out of the event-driven nature of the sensor output, they cannot  fully utilize its advantages: the clocked, sequential operation of a CPU, as well as separation between memory and processor stay in contrast to the highly parallel asynchronous temporal stream of events coming from an event-based sensor. 

Neuromorphic hardware, in contrast to the conventional CPU,  offers a massively parallel computing substrate that is inherently event-based, which matches the processing paradigm of event-based sensors \cite{Indiveri2015,Davies2018,MerollaEtAl2014,FurberEtAl2012}. We aim to develop neuronal architectures that solve different tasks on such neuromorphic devices taking full advantage of the event-driven computation. In this work, we target a vision task of object tracking; in particular we show how a simple attention network can be configured on a neuromorphic hardware to select one object in an event-based input stream and to track this object in presence of equally salient distractors. While similar principles to the one realized here have been used in early days of neuromorphic engineering to design a dedicated spiking neuromorphic chip for attention \cite{Indiveri2000}, here we present its  realization on a generic neuromorphic device Loihi  that can also support other vision, cognitive, and motor control tasks \cite{Davies2018}.

Neuromorphic processors emulate dynamics of biological spiking neurons in hardware and thus allow us to run spiking neural network architectures in real-time, with a small energy footprint, and in small form-factor devices, making them a promising computing platform for event-based vision~\cite{Indiveri2015,Davies2018}. However, most neuromorphic devices target offline computation, with applications in either computational neuroscience \cite{FurberEtAl2012,Aamir2018} or data processing \cite{MerollaEtAl2014}. The Kapoho Bay -- a USB-stick form-factor version of the Intel's latest neuromorphic research platform Loihi \cite{Davies2018} -- is among neuromorphic systems that offer a direct Address-Event Representation (AER) interface to event-based sensors \cite{Boahen99}. This allows us to build a setup, in which a neuromorphic camera DAVIS \cite{Brandli2014} can be directly interfaced to Loihi, stimulating on-chip neurons configured in a network that can solve vision tasks. Here, we focus in particular on the task of object-centered attention and tracking. 

 While the event-based output of a DAVIS camera singles out a fast-moving object easily, two capabilities  need additional processing: the ability to suppress distractors even if their salience changes and at times surpasses that of the target object and; keeping track of an object if it slows down or even stops. To gain these capabilities, the system needs a mechanism to hold a memory of the object's location in the field of view. While such a memory mechanism can be realized on a conventional computing system, doing so would alleviate the advantages of the low-power event-based computing. Here we explore a setup in which an event-based sensor is interfaced to an event-based processor running a recurrent neural network that is capable of creating memory states based on the incoming events.

Feed-forward artificial neural networks (ANNs) are stateless -- they merely transfer inputs to outputs and if they do not receive input, the activity in the network fades away. In a spike-based network with integrate-and-fire neurons, the activity decays with a time-constant of neuronal dynamics, in a conventional ANN, even more radically, on the next clock cycle. In order to create a memory state, recurrence in the network is needed. A well-known model for working memory that has been studied in computational neuroscience and cognitive science is a Dynamic Neural Field (DNF) \cite{SchonerSpencer2015} -- a neural population-based model. The DNF is a dynamical system that can be realized as a recurrent neural network with attractor dynamics, created by configuring a population of neurons with a winner-take-all  connectivity~\cite{Sandamirskaya2014}. In this connectivity pattern, neurons that encode similar values have an excitatory connection and neurons that encode different values -- an inhibitory one. This simple connectivity pattern performs a selective amplification of a noisy input and, in an extreme case of strong interaction, can create sustained activation patterns that are kept active even if the initial input ceases completely. This property has been used as a model of working memory \cite{SchonerSpencer2015}.

DNFs have been used previously to realize object tracking with on SCAMP -- a smart camera with an in-focal- plane processor array \cite{MartelSandamirskaya2016}. Here, we demonstrate object tracking with DNFs in an event-based setting using a spiking neuromorphic device Loihi.

\section{The hardware setup} 

\subsection{Neuromorphic Device Loihi}

Intel Neuromorphic Computing Lab designed the neuromorphic research chip Loihi, in which spiking neural network models can be simulated in real-time efficiently \cite{Davies2018}. The chip consists of a mesh of 128 neuromorphic cores, three embedded x86 processor cores, and an off-chip communication interface that allows to scale up architectures to multiple Loihi devices.  Compartments are the main building blocks used to configure both single- and multi-compartment neurons. In this work we only use single-compartment neurons.

The external input to a network on Loihi is provided through spike generators. Spike generators are ports connected to compartments that can generate spikes at precise time-steps. Loihi provides an instrument for measuring the variables and sending them  off the chip using ``probes''. For compartments it is possible to define probes to measure spike events, neuron's membrane voltage and input current. For the connections, probes can measure multiple synaptic variables, including weight, pre-synaptic and post-synaptic traces. Probing, however, affects the performance of the chip.

Loihi's Python NxSDK-0.8.0 API allows us to implement SNNs on the chip~\cite{Lin2018}. The NxNet API provides ways to define a graph of neurons and synapses and configure their parameters (such as decay time constants, spike impulse values, synaptic weights, refractory delays, spiking thresholds), inject external stimulus into the network, implement custom learning rules, and monitor and modify the network during runtime. 

\subsection{Dynamic Vision Sensor DAVIS}

In this work, we used a Dynamic Vision Sensor type of a camera, the DAVIS240C~\cite{Brandli2014}. The DAVIS camera emulates the dynamics of biological retinal cells in silicon using mixed-signal analog/digital technologies. 
There are $240\times180$ pixels integrated on the chip. 
Each pixel independently detects the brightness  change in a small area of the visual scene and emits an event if the brightness change passes a positive (``on'' event) or a negative (``off'' event) threshold. Each event is a digital data packet that carries the address of the pixel, the polarity of the detected brightness change, and the time of the event using Address Event Representation. Due to its high dynamic range, the sensor captures moving objects in its visual field in a wide range of lighting conditions. 

To connect DAVIS to Loihi, the direct parallel AER interface on the device can be used. The events are captured and distributed to neurons on the neuromorphic cores through the embedded FPGA and x86 processors. For reproducibility, in this work, we use recorded spikes that we feed into Loihi using a spike generator from the NxNet API to generate some of the plots. The network was also tested with the direct AER interface between the DAVIS and Loihi.

\section{Attractor Dynamics for Object Tracking: the Dynamic Neural Fields}

\begin{figure}
\includegraphics[scale = 0.70]{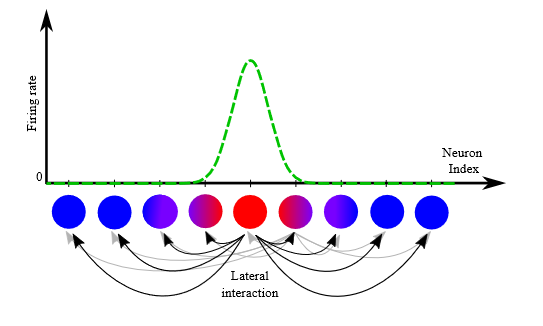}
\caption{Schematic representation of a 1-dimensional winner-take-all dynamic neural field. The lateral connections are all-to-all and the synaptic weights are defined by the kernel function that depends on the distance between the pre- and post-synaptic neurons.}
\vspace{-0.3cm}
\label{WTA_1D} 
\end{figure}

A Dynamic Neural Field is a mathematical model that was derived to describe activity of large homogeneous populations of biological neurons~\cite{SchonerSpencer2015}. The connectivity pattern in such a neuronal population is shown in Fig.~\ref{WTA_1D}. First, neurons are ``aligned'' according to a feature which they are sensitive to (here, the position in the camera's field of view). Second, neurons that are sensitive to similar values of the feature (i.e. are close to each other on this behavioral space) are connected via excitatory (positive weight) connections, and neurons that are sensitive to dissimilar features inhibit each other (negative-weight connections). This soft winner-take-all connectivity pattern, combined with a non-linearity of the neuronal activation function, leads to formation of an attractor state in a DNF neuronal population. In particular, DNFs form a so-called bump-attractor -- a localized activity peak centred over a salient value of the behavioral variable (green line in Fig.~\ref{WTA_1D}). Such bump-attractor networks have been used in the past both to explain activity patterns in biological neural networks and to build artificial cognitive systems for robot control \cite{SandamirskayaEtAl2013,SchonerSpencer2015}.
To realize DNF dynamics on Loihi, we create a group of compartments that are connected with synapses with weights matching the "Mexican-hat" connectivity kernel.

The output of the DNF population is computed as a population vector using the instantaneous firing rate of neurons that is inversely proportional to the inter-spike intervals.

%

\section{Results}

\subsection{Attractor dynamics on a neuromorphic Chip}

\begin{figure*}
\centering
\subfigure[\label{fig:select_rand}]{
\includegraphics[width = 0.48\textwidth]{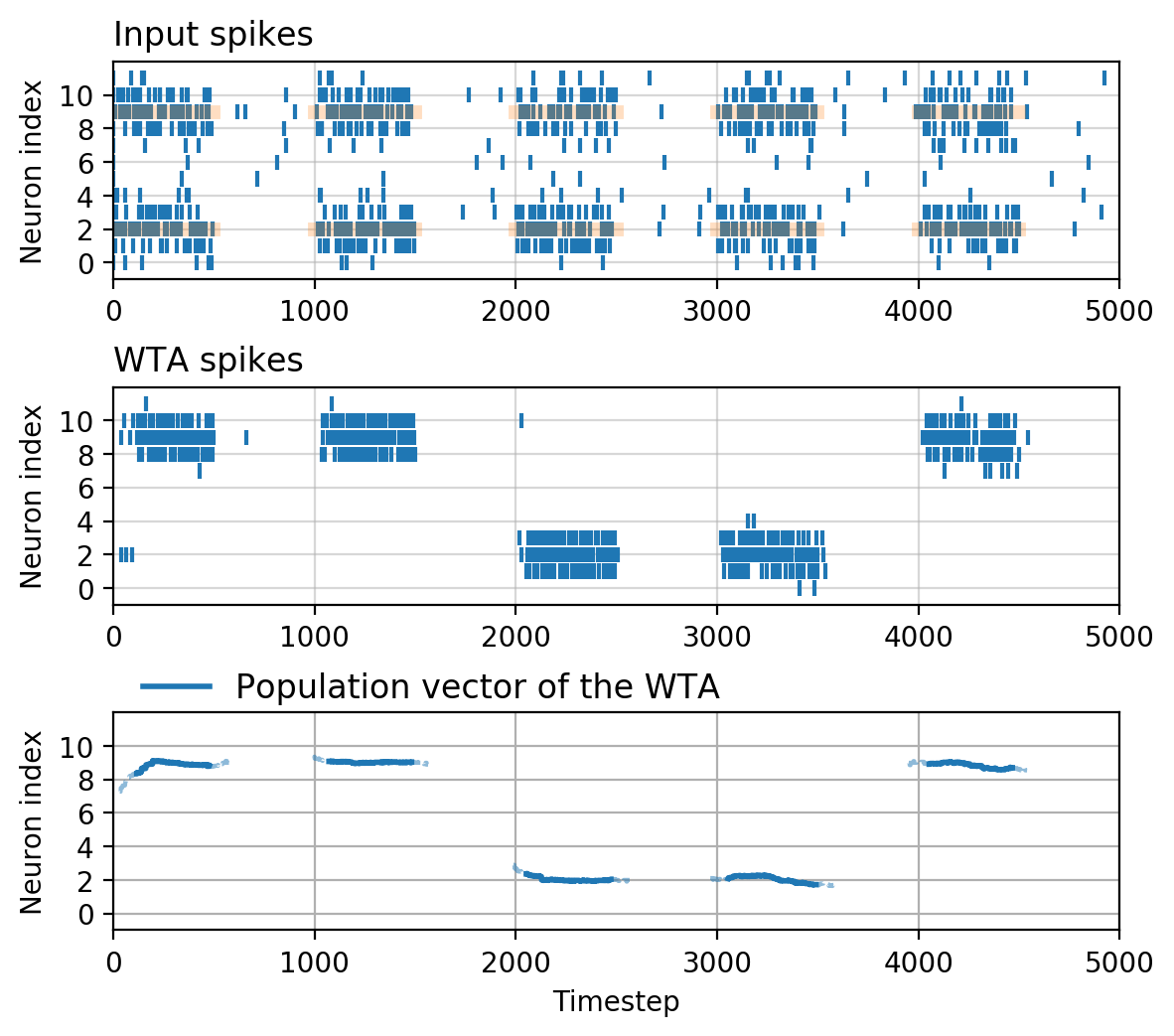}}
\subfigure[\label{fig:select}]{
\includegraphics[width = 0.48\textwidth]{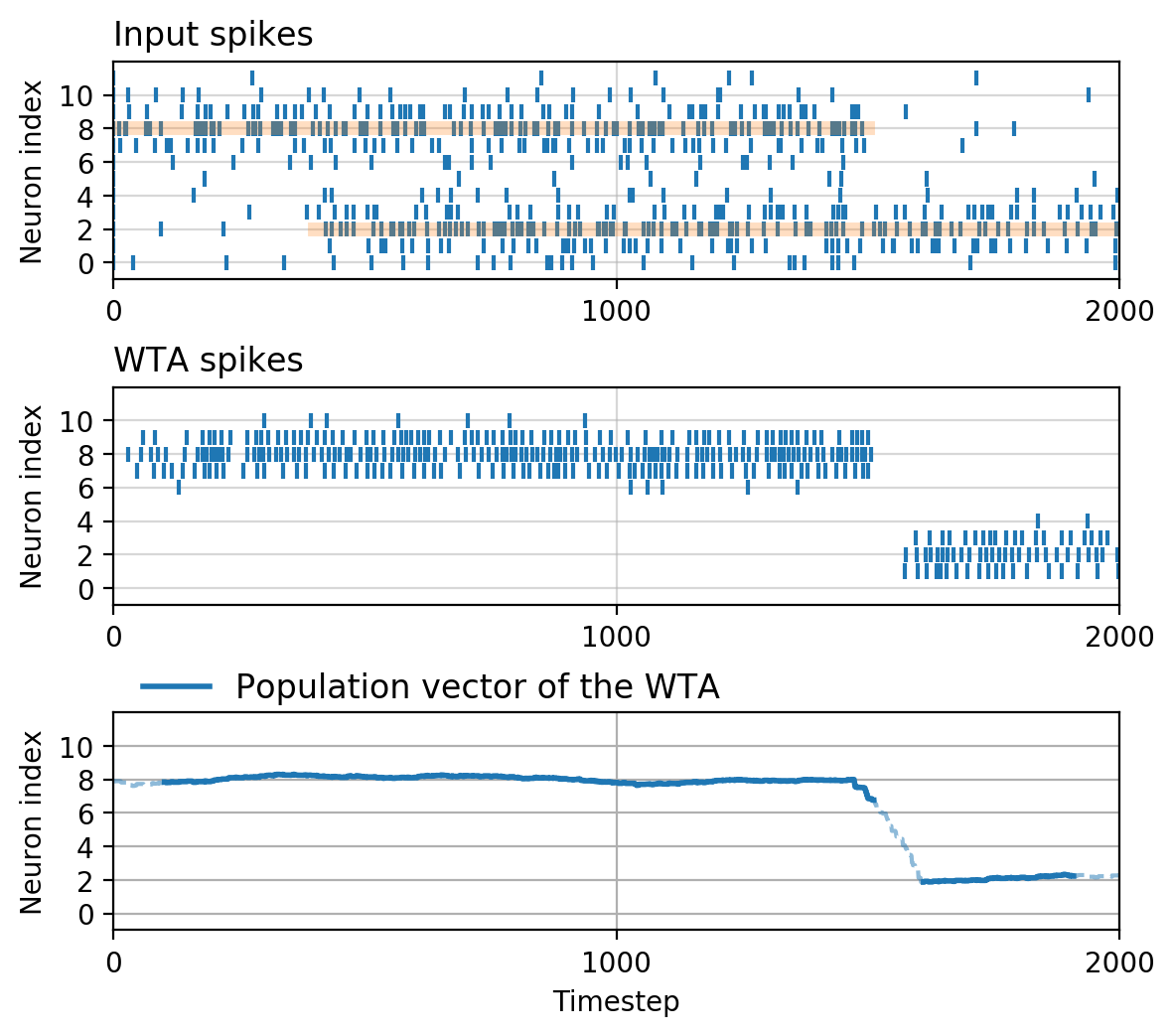}}\\
\subfigure[\label{fig:nonselfsust}]{
        \includegraphics[width = 0.48\textwidth]{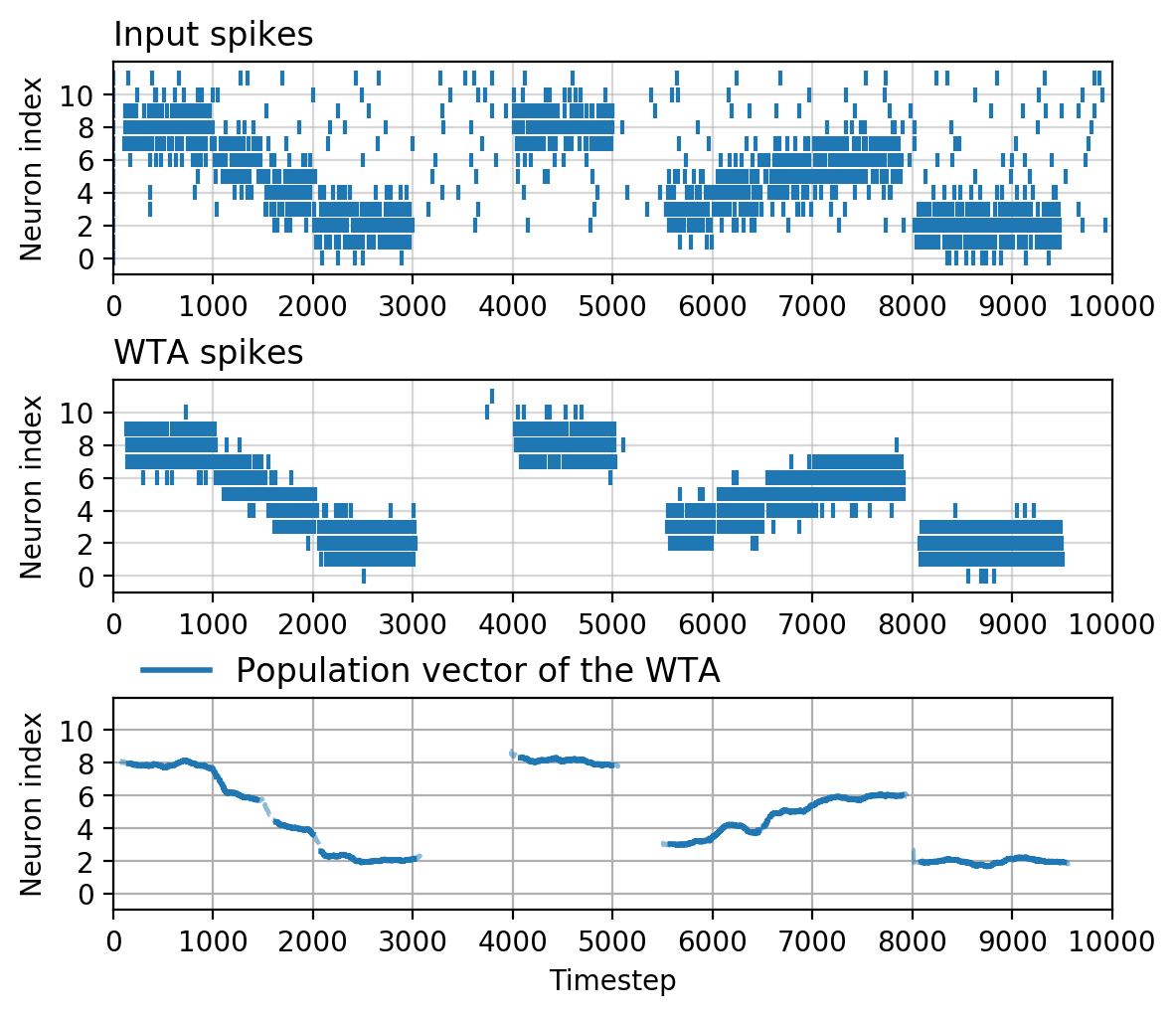}}
\subfigure[\label{fig:selfsust}]{   
    \includegraphics[width = 0.48\textwidth]{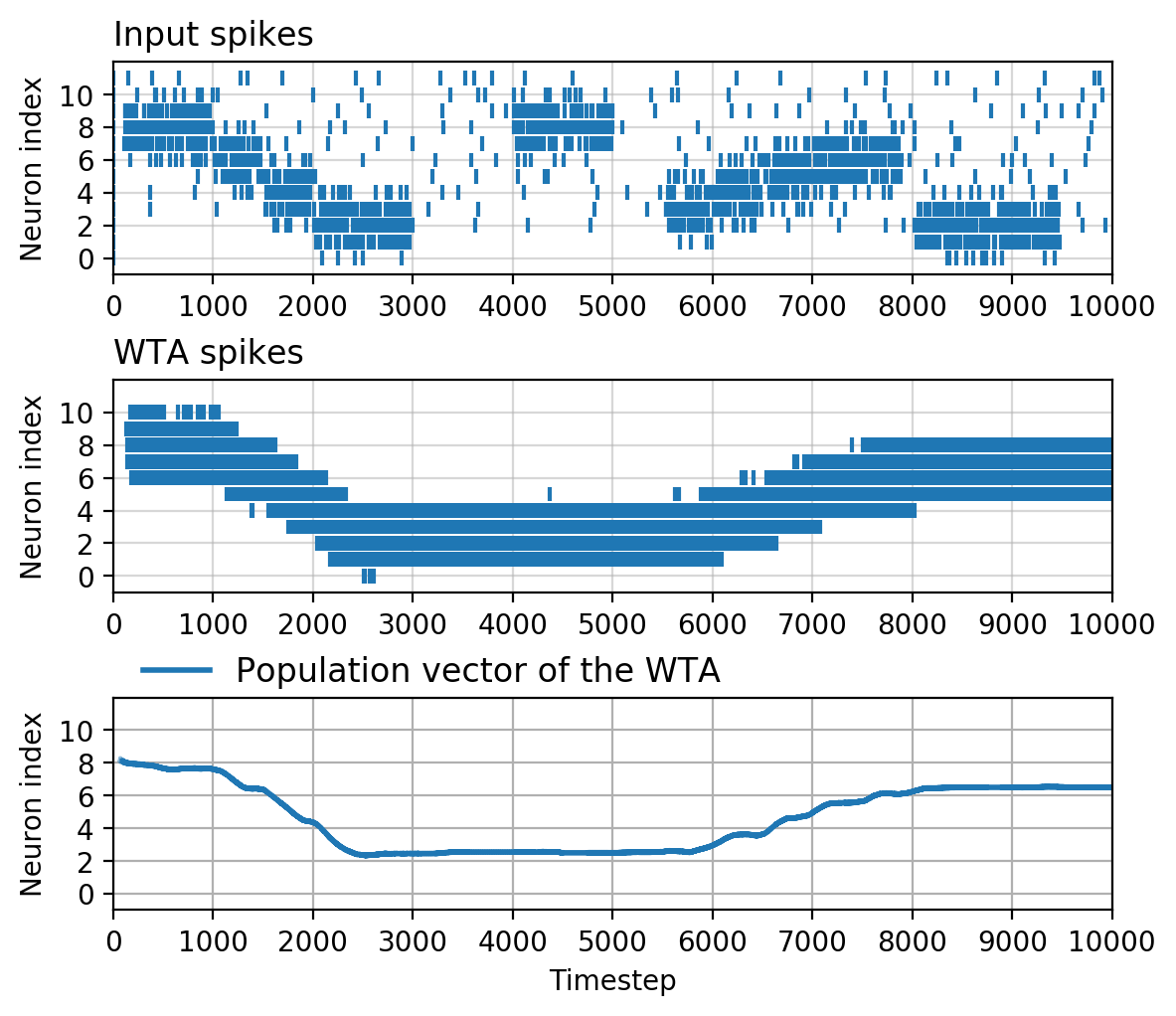}}
\caption{The plots show activity of spiking neurons on the Loihi chip, configured as a DNF in an input-driven (a, b, c) and self-sustained (d) regime. In each subfigure, the top plot shows input spikes generated on the computer, the middle plot shows   output of the DNF population on Loihi, and the bottom plot shows the population vector (mean of the activity) of the DNF population.
}
\label{wta_selective_1}
\end{figure*}

First, we demonstrate the properties of a DNF realized in a spiking neural network on Loihi that are used in the object-tracking application. We have configured a small population of 12 neurons in a winner-take-all fashion, as shown in Fig.~\ref{WTA_1D}. We used two sets of parameters to demonstrate two dynamical configurations of the DNF: a selective input-driven regime, shown in Fig.~\ref{wta_selective_1}a, b, c and a self-sustained regime shown in Fig.~\ref{wta_selective_1}d. In each subfigure, the upper plot shows spikes from the spike generators that send input to Loihi, the middle plot shows the output spikes from the DNF population, and the lower plot shows population activity vector of the DNF population (the position of the mean of neuronal activity) over time of the experiment. 

Fig.~\ref{wta_selective_1}a shows that a DNF configured in a selective regime selects one of the input bumps in the case of a bi-modal input distribution. For each pair of input bumps with equal strength (average firing rate), the DNF selects one of them randomly. Fig.~\ref{wta_selective_1}b shows behavior of the same DNF population for an input sequence, in which one of the input bumps arrives first, is selected and then stabilized by the lateral interactions in the neuronal population. In this configuration, the second input bump is rejected by the DNF population and does not lead to any activity of the neurons in the respective region.

\begin{figure}
\centering
\includegraphics[width = 0.39\textwidth]{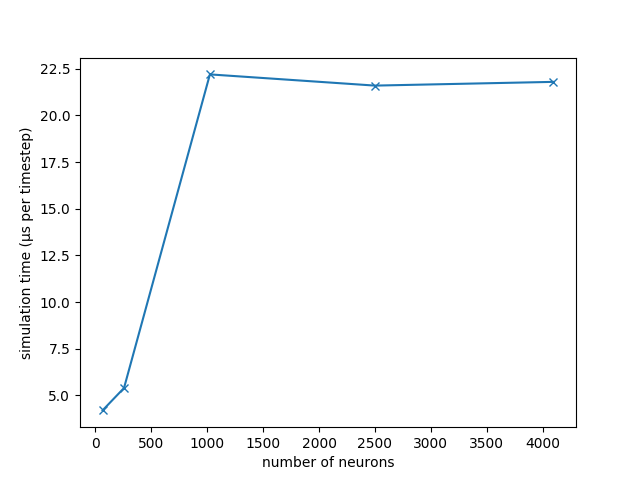}
\caption{Performance evaluation. }
\vspace{-0.5cm}
\label{fig:performance}
\end{figure}

Fig.~\ref{wta_selective_1}c and d contrast behavior of the DNF configured to be in an input-driven regime (c) versus in a self-sustained regime (d). For the same input, the input-driven DNF (Fig.~\ref{wta_selective_1}c) follows the input activity, only rejecting noise around the activity bumps. The self-sustained DNF (Fig.~\ref{wta_selective_1}d) keeps the position of the selected object and ignores input at different locations unless it is spatially proximal to the current activity bump. Thus, it shows tracking behavior.

To configure the two DNFs, we used the parameters on Loihi listed in
the Table~\ref{table_1}. In particular, we use Gaussian-shaped connectivity profile for lateral connections in the WTA (of amplitude ``Excitatory weight'' and spread ``Connectivity kernel $\sigma$'') and a direct global inhibition.   Input- and background-noise are generated as Poisson spikes.

 \begin{table}
 \centering
 \begin{tabular}{ |p{4cm} | p{1.5cm} | p{1.5cm}| }
 \hline
  Parameter & Input driven & Self-sustained \\
 \thickhline
 Voltage threshold   & $3000 * 2^6$    &$3000*2^6$\\
 Voltage decay time constant &150 ts & 150 ts\\
  Current decay time constant  &10 ts & 10 ts\\
  Connectivity kernel $\sigma$ & 1.5  & 1.5 \\
  Self excitation & no & yes \\
  Excitatory weight& 200  & 150 \\
  Global inhibitory weight & -160 & -75 \\
  Input weight & 200 & 200 \\
  Input firing rate (Poisson)& 60 Hz & 60 Hz \\
  Noise firing rate (Poisson)& 2 Hz & 2 Hz \\
 \hline

 \end{tabular}
 \caption{DNF parameters used to produce plots in  Fig.~\ref{wta_selective_1}}
 \label{table_1}
 \end{table} 

\subsection{Performance evaluation}

Fig.~\ref{fig:performance} shows the  duration of simulation on Loihi per time step depending on the number of neurons in a 2D DNF population with a single self-sustained bump without spike generators (apart from the initial one creating the bump) or probing. Neurons are distributed over the 2x128 cores. One can observe that the simulation time per time step stays on the level of 20$\mu s$ for network sizes above 1000 neurons\footnote{Although quite fast already, the simulation time depends on the details of the network implementation and can be further optimized.}.

\subsection{Event-based attention and tracking}

\begin{figure}
\centering
\includegraphics[width = 0.45\textwidth]{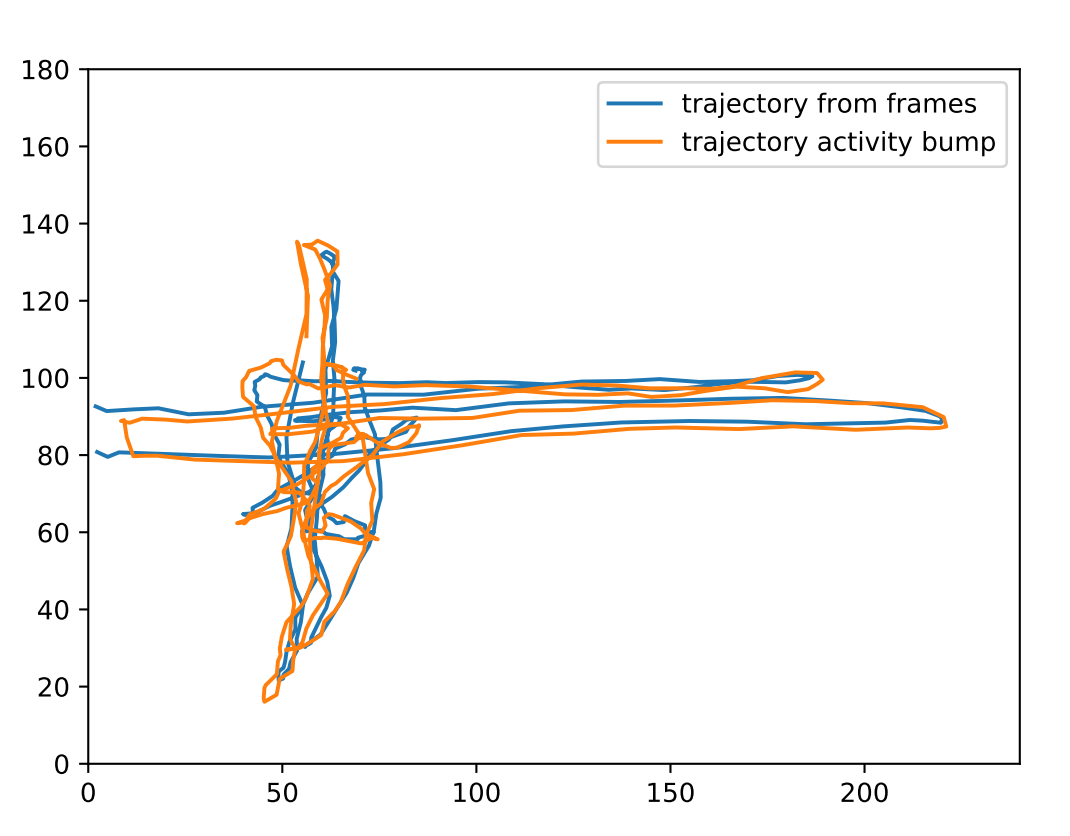}
\caption{
Trajectory of the selected object (star), obtained from the activity bump in the DNF on Loihi and from frames, used as ground truth.  The average distance between the points in the two trajectories over all frames is 3.5 DAVIS pixels. The distance was calculated with an offset between frames- and events-trajectory of 15 ms, where the distance is minimal.
}
\label{fig:trajectory}
\end{figure}

The tracking network consists of two two-dimensional WTA/DNF layers with 64x64 neurons each. The first layer has excitatory connections at a level at which peaks  are self sustained and  weak global inhibition so that multiple bumps can form. Bumps form at the locations of the on-event input and follow the on-events when objects move over the field of view of the DVS. The off-events inhibit neurons in this population, decreasing activity in the bumps. This facilitates fast moving bumps avoiding a tail in the activation pattern. The second WTA layer receives input from the first layer through  one-to-one connections. This layer has strong self-excitation and strong global inhibition which lead to the selection of a single bump. To this layer, we provide an excitatory  initial input, cuing one of the objects in the beginning of the DVS stimulation; feature-based cues can also be used here~\cite{SchonerSpencer2015} (Chapter 5). The WTA forms an activity bump over the selected object, which is moved by the excitatory input from the input layer when the selected object moves. In these experiments, parameters listed in Table~\ref{table_2} were used. In particular, both DNF layers are self-connected  using a Mexican hat connectivity kernel specified by a difference of two Gaussians (with amplitude and $\sigma$ of the excitatory and inhibitory kernels); a global inhibitory group of neurons is used to provide global inhibition.  Every neuron in the inhibitory group has a given probability to be connected to any neuron in the excitatory layer.

 \begin{table}
 \centering
 \begin{tabular}{ | p{6cm} | p{1.2cm} | p{0.0cm} |  }
  \hline
  Parameter & value \\
  \thickhline
 Voltage threshold   & $640 * 2^6$   \\
 Voltage threshold global inhibition   & $896 * 2^6$   \\
 Voltage decay time constant &20 ts \\
 Current decay time constant &20 ts \\
 Input connectivity kernel $\sigma$  & 1.5 \\
 Excitation kernel  $\sigma$  & 2 \\
 Inhibition kernel  $\sigma$  & 4 \\
 Excitatory weight non-selective & 152  \\
 Inhibitory weight non-selective & -41  \\
 Excitatory weight selective & 230  \\
 Inhibitory weight selective & -41  \\
 Excitatory weight to global inhibition & 5 \\
 Global inhibitory weight non-selective & -20  \\
 Global inhibitory weight selective & -90  \\
 Excitatory weight non-selective to selective (1:1 connectivity) & 740  \\
 Input weight on events & 70  \\
 Input weight off events & -50  \\
 Refractory period & 12 ts\\
 Refractory period global inhibition & 7 ts\\
 Connection probability to/from global inhibition & 0.6\\
 Number of global inhibitory neurons  & 40\\
 \hline
 \end{tabular}
 \caption{DNF parameters used to produce plots in  Fig.~\ref{fig:tracking} and \ref{fig:tracking2}, if layer is not specified, the parameter applies to both.
 }
 \label{table_2}
 \end{table}

\begin{figure*}
\centering
\subfigure[\label{fig:frames}]{
\includegraphics[width = 0.9\textwidth]{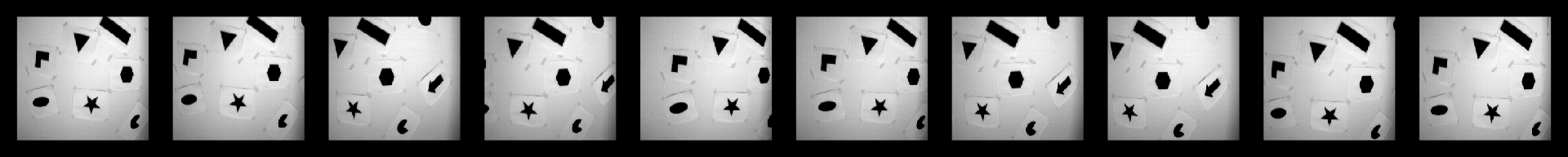}} 
\\
\subfigure[\label{fig:inp}]{
\includegraphics[width = 0.9\textwidth]{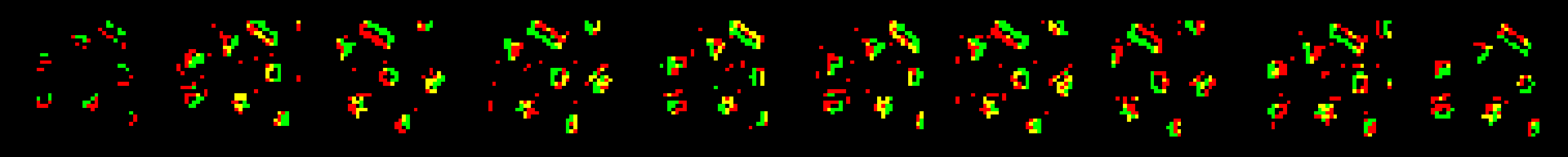}}
\\
\subfigure[\label{fig:exc}]{
\includegraphics[width = 0.9\textwidth]{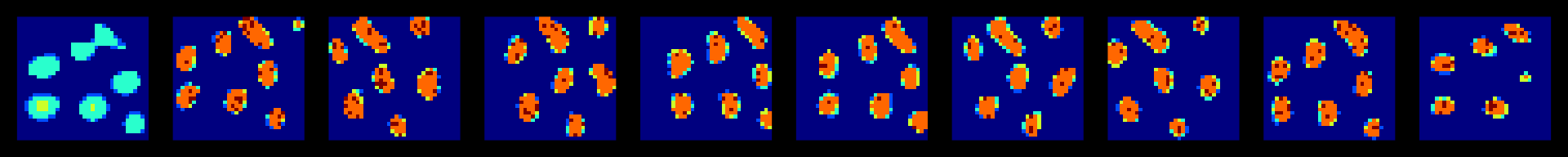}}
\\
\subfigure[\label{fig:out}]{
\includegraphics[width = 0.9\textwidth]{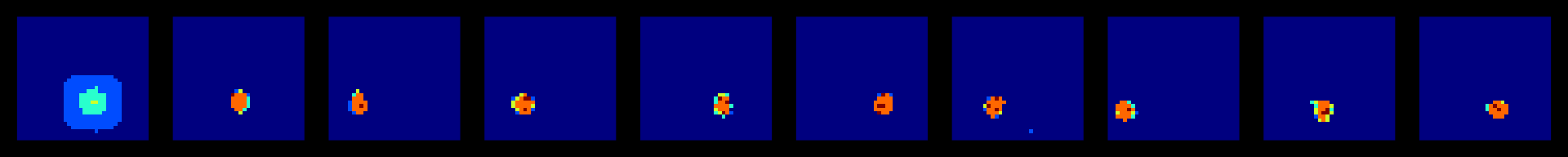}}
\caption{
Object tracking experiment: (a) snapshots of input DAVIS frames (top); (b) DAVIS on (green) and off (red) events binned into 10ms frame (second from top); (c) Firing rate of first non-selective WTA layer on Loihi (third from top); and (d) second, selective WTA layer on Loihi (bottom).
}
\label{fig:tracking}
\end{figure*}

Fig.~\ref{fig:tracking} shows performance of the two-dimensional DNF on the Loihi chip in a tracking experiment. Here, the shapes\_translation dataset \cite{Mueggler2017} is used that contains a number of objects drawn on a wall in front of a moving DVS. Plots in Fig.~\ref{fig:tracking}b show input that is sent to Loihi over the course of the experiment: the DVS events are emitted at the edges of the objects. 

Fig.~\ref{fig:tracking}c shows activity of the first, non-selective DNF layer: activity in this layer forms peaks over all objects, stabilizing this activity in moments with reduced motion and weaker DVS output. Fig.~\ref{fig:tracking}d shows activity of the output layer of the tracking DNF. Here, one of the objects (the star) is selected (by a local boost to this layer in the beginning of the simulation) and is tracked throughout the experiment, despite presence of the distractors. 

To obtain the spike plots, the spikes were filtered with a 50ms rectangular filter and snapshots were taken at regular intervals within the simulation of 6500 time steps. DAVIS input events to the first layer were down-sampled and binned into 1ms per time step (events that exceeded one event per bin, were discarded), i.e. the 6500 time steps correspond to 6.5s of DAVIS input.

Fig.~\ref{fig:trajectory} shows the trajectory of the selected object that is extracted from the activity of the output layer of the DNF model and the ground truth extracted from the input frames.  The blue trace shows the center  of the star object extracted from the DAVIS frames by thresholding (ground truth). The orange trace shows mean (i.e. population vector) of the instantaneous firing rate of neurons  in the second (output) layer of the network (i.e. the middle of the tracked bump). Instantaneous firing rates are estimated based on the inter-spike intervals.  

The DVS input to the network was down-sampled to 64x64 neurons, the trajectory was up-sampled to the DAVIS resolution of 240x180. The mean error was calculated as the mean of all distances between the locations of the bump activity and the locations of the frame based extraction at the timesteps of the DAVIS frames and amounts to 3.5 DAVIS pixels.\footnote{The Figure was generated using a Brian2 simulation of the equations implemented in Loihi, as it is currently impossible to probe a large network for that many time steps, however, performance of the network can be observed in a live demo.}

\begin{figure*}
\centering
\subfigure[\label{fig:inp_2}]{
\includegraphics[width = 0.9\textwidth]{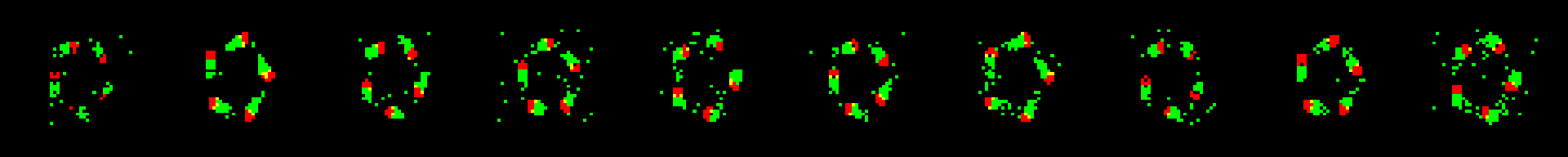}} \\
\subfigure[\label{fig:exc2}]{
\includegraphics[width = 0.9\textwidth]{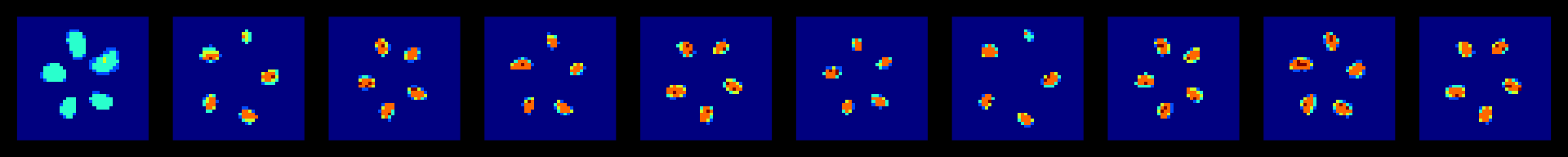}} \\
\subfigure[\label{fig:out2}]{
\includegraphics[width = 0.9\textwidth]{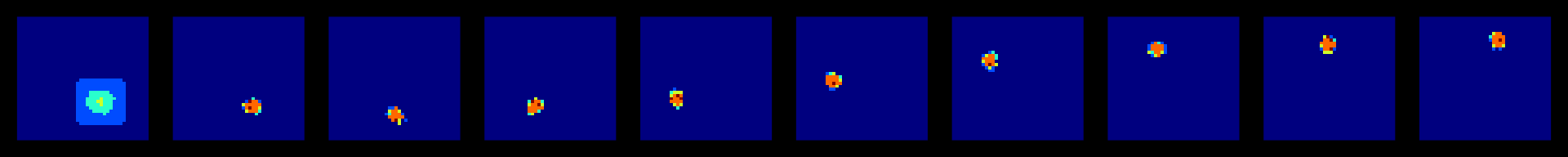}}
\caption{
Object tracking experiment 2: DAVIS on (green) and off (red) events binned into 10ms frames (top); firing rate of the first non-selective WTA layer on Loihi (middle);  output of the second, selective WTA layer on Loihi (bottom).
}
\label{fig:tracking2}
\end{figure*}

Fig.~\ref{fig:tracking2} shows our second tracking experiment. Here, a ring with five identical objects is rotating in front of the DAVIS camera. The first layer of the tracking SNN architecture creates activity bumps for all five objects, while the second layer (bottom plots) only tracks the single object, selected by a localized activity boost in the beginning of the experiment (first pane in the plot). The same parameters were used here as for Fig.~\ref{fig:tracking}. The length of simulation on Loihi was 3000 timesteps here. DAVIS input events were binned into 0.5ms per timestep, i.e. the simulation corresponds to 1.5s of DAVIS input.

\section{Conclusion}

In this work, we have shown for the first time a setup that interfaces an event-based camera DAVIS with the spiking neuromorphic system Loihi, creating a purely event-driven sensing and processing system. We have implemented a simple attention and tracking network on Loihi that allows to select a single object out of a number of moving objects in the visual field and track this object, even in cases when movement stops and the event stream is interrupted. Full evaluation of the system in terms of tracking speed and quality, as well as power efficiency and robustness is target of our current work and will be reported shortly, while functioning of the system can be observed in a live demonstration during the workshop.  

\section*{Acknowledgements} We would like to thank Dr. Julien Martel and the Intel Neuromorphic Computing Lab for their help with the hardware and software setup used in this work.  This project has received funding from the SNSF project Ambizione (PZ00P2\_168183) and a ZNZ Fellowship from the Neuroscience Center Zurich. 
\\ 

{\small
\bibliographystyle{IEEEtran}
\bibliography{literature_mendeley,my_bib}
}

\end{document}